\documentclass[runningheads]{llncs}

\usepackage[T1]{fontenc}
\usepackage{graphicx}
\usepackage{amsmath,amssymb,amsfonts}
\usepackage{textcomp}
\usepackage{xcolor}
\usepackage{tikz}
\usepackage{booktabs}
\usepackage{tabularx}
\usepackage{float}
\usepackage{placeins}
\usetikzlibrary{positioning}
\usepackage{bbding}

\usepackage{float}
\usepackage{placeins}
\begin{document}

\title{Compression-Based Behavioral Similarity for Open-World Sybil Discovery on Ethereum}

\author{Michał Bartnicki\inst{1}\orcidID{0009-0004-5856-4850}\Envelope \and
Jarosław A. Chudziak\inst{1}\orcidID{0000-0003-4534-8652}}

\titlerunning{Sybil Discovery on Ethereum}
\authorrunning{Bartnicki, Chudziak}

\institute{
Faculty of Electronics and Information Technology,
Warsaw University of Technology, Warsaw, Poland\\
\email{michal.bartnicki.stud@pw.edu.pl}\\
\email{jaroslaw.chudziak@pw.edu.pl}
}

\maketitle

\begin{abstract}
Sybil attackers are Blockchain actors that adopt the characteristics of regular users to exploit airdrops or influence governance.
Current methods of Sybil actor detection include constructing graphs, which requires token transfers between examined wallets. Machine learning algorithms have been employed as well, but they treat the task as a closed-set classification problem, making them vulnerable to frequent changes in attack strategies or evasion tactics.
We address the following questions: can compression-based similarity differentiate Sybil bots, organic users, and arbitrage bot wallets without direct financial links? What is the effect of high-signal contracts on the discovery of Sybils, and how robust are behavioral graphs under temporal drift and adversarial perturbations?
Our approach synthesizes a symbolic Transaction Grammar from EVM (Ethereum Virtual Machine) traces, capturing separately transaction rhythm, execution structure, and functional intent. The high-signal contracts are filtered with our own protocol, called the Blind-Spot Protocol. Gzip-based NCD is used to construct a behavioral graph for Sybil discovery. We validate this framework against supervised machine learning baselines, a temporal split, and synthetic camouflage stress tests.
Ultimately, we contribute a leakage-aware behavioral framework for Sybil candidate discovery. Its core NCD primitive requires no supervised training and can expand suspicious seed wallets without explicit funding links. We position the method as a training-free local discovery primitive for open-world blockchain audits, rather than as a formal open-set recognition system.

\keywords{Blockchain Analytics \and Ethereum \and Sybil Discovery \and Normalized Compression Distance \and Behavioral Similarity }
\end{abstract}

\section{Introduction}
\label{sec:introduction}

With DeFi and governance growth in Ethereum, Sybil attacks remain a persistent threat \cite{thesybilattack}. In these attacks, one entity controls many pseudonymous wallets to farm airdrops or manipulate DAO governance \cite{zhou2023sokdecentralizedfinancedefi,liu2025detectingsybiladdressesblockchain,messias2025airdropsgivingmoneyaway}. Aggregate statistics or financial links expose simple cases, but evasive campaigns require coordination and execution-pattern analysis.

Current detection strategies primarily rely on explicit financial interaction graphs and closed-set supervised classifiers \cite{Alipanahloo_2024}. However, this framing is poorly aligned with blockchain audit settings where labels are incomplete, strategies evolve after deployment, and the goal is often candidate discovery rather than closed-set classification. \cite{TowardOpenSetRecognition}. Graph-based methods attempt to detect communities with common funding or value flow \cite{AddressClusteringHeuristics}, although such indicators become weaker when value flows pass through centralized exchanges, disperser contracts, or other intermediates. Tabular feature-based supervised methods, including balance, transaction rate, gas expenditure, and number of interactions, have shown promise if representative labels exist \cite{huang2022ethclass,liu2022fagnn}. Recent benchmarks suggest tree-based models remain the gold standard for such tabular blockchain data \cite{grinsztajn2022treebasedmodelsoutperformdeep,shwartzziv2021tabulardatadeeplearning}. But such methods suffer from two weaknesses in audit-driven candidate discovery. First, models may exploit shortcut features, such as label-indicative contracts, instead of transferable behavior. The problem has parallels to more general issues around robustness of ML systems against adversarial distributional shortcuts \cite{kan2025tesseracteliminatingexperimentalbias}. Second, they remain reactive until representative labels are curated.

In this paper, we posit that Sybil discovery in this audit setting can be rephrased as pairwise behavioral similarity analysis rather than categorization of individual wallets. We define a wallet's behavior using a symbolic \textit{Transaction Grammar} that encodes transaction rhythm, EVM execution trace structure, and function-level intent. Modeling wallet behavior as a sequence lets us measure similarity using Algorithmic Information Theory without training a classifier.

\section{Related Work}
\label{sec:related_work}

The process of Sybil detection generally involves the use of explicit graphs, where each edge denotes an ETH or ERC-20 token transfer \cite{UnderstandingEthereumviaGraphAnalysis}. Community detection and clustering methods are used to detect Sybils with similar funding channels, whereas GNN pipelines classify Sybil accounts based on graph-based features \cite{liu2022fagnn,AddressClusteringHeuristics}. Although such techniques are effective, their performance depends heavily on visible funding flows and is significantly reduced when actors hide token transfers through intermediaries such as exchanges or dispersers.

The second method uses tabular machine learning to detect blockchain actors. Such algorithms operate with aggregate wallet statistics like number of transactions, transaction timings, gas usage, balance changes, and interaction frequencies \cite{huang2022ethclass,Niedermayer_2024,flashboys}. The works on financial bot detection on Ethereum also confirm the efficiency of supervised and unsupervised machine learning approaches provided suitable features and labels are available. In particular, recent publications construct labeled Ethereum human/bot datasets using timing, frequency, gas price, and gas-limit features.

The current work on blockchain analytics seeks to leverage more advanced behavioral representations, taking into account that bots are characterized by their rhythm of activity and patterns of interaction, not the sheer rate of communication. Existing learning-based approaches have mostly been supervised and task-specific, labeling predefined classes of wallets, but failing to uncover new bot communities. This makes them inadequate for open-world blockchain audits, where new bot scripts may emerge before representative labels are available. Evaluations suffer from similar problems related to adversarial learning. One example is provided by TESSERACT, which highlights how incorrect experimental setups can cause an overestimation of performance because of space-time bias \cite{kan2025tesseracteliminatingexperimentalbias}. In the blockchain setting, a comparable pitfall is when learning algorithms take advantage of high-signal counterparts or blockchain peculiarities rather than extracting universal behavioral patterns.

Our approach is rooted in the methodological foundation of Normalized Compression Distance (NCD). The NCD evaluates similarity based on the compressibility of an artifact when compared to its counterpart; hence, the similarity of an object is measured depending on how well it can be compressed together with another artifact. Compression has been used before for various other purposes such as malware classification and cybersecurity; in all of those cases, similar structures of two different objects could have been identified despite their dissimilarity \cite{cilibrasi2004clusteringcompression}. So far, there is no research related to compression and similarity of Ethereum Sybils.

In other words, our technique can be positioned between two well-established frameworks. First, contrary to explicit graph approaches, our method does not rely on explicit financial connections between wallets. Second, unlike supervised classifiers, no labeled training is needed in order to compute similarity scores between wallets. Instead, we define implicit graph where edges denote similarity of structural properties of transaction grammars. Such a construction allows local candidate generation from a suspicious seed wallet. Our approach targets audit scenarios where labels are rare, actor behaviors evolve, and transaction connections may be unavailable. Accordingly, graph neural baselines over explicit transfer or funding graphs are complementary rather than directly comparable to the disconnected behavioral-similarity task evaluated here.

\section{Problem and Approach}
\label{sec:method}
This section presents our behavioral formulation of open-world Sybil discovery. We define the task as candidate discovery over wallet behavior, encode wallet histories as Transaction Grammars, compare them with Normalized Compression Distance, and construct an implicit behavioral graph. We then introduce the Blind-Spot protocol used to control leakage from high-signal contracts before evaluation.
\subsection{Problem Formulation}
\label{sec:problem_formulation}

Sybil attackers threaten blockchain ecosystems by operating many pseudonymous wallets that imitate ordinary users to farm airdrops or influence decentralized governance. The task is not simply bot detection: arbitrage-oriented MEV bots are also automated, but may represent benign or economically motivated activity. We therefore distinguish coordinated Sybil behavior from both Organic users and MEV automation.

Explicit financial graphs are effective when wallets share funding or transfer links, but coordinated campaigns may appear disconnected after using exchanges, dispersers, or intermediaries. Supervised classifiers require curated labels, retraining, and may exploit shortcut features such as class-indicative contracts. We instead formulate Sybil detection as training-free candidate discovery in an open-world audit setting: given wallet histories, identify groups that share behavioral grammar without requiring direct financial links or a closed-set classifier.

Each wallet \(W\) is represented as an ordered sequence \(S_W=\{x_1,x_2,\dots,x_L\}\), where each transaction \(x_t\) encodes rhythmic, structural, and intent-based properties from EVM traces. NCD then measures shared regularity between sequences and induces an implicit behavioral graph whose edges reflect script-level similarity rather than token flow.

This formulation leads to four empirical questions: whether Sybil wallets are internally more similar than Organic and MEV wallets; whether compression-based similarity supports candidate discovery without financial links; how much signal comes from high-signal contract leakage; and how stable the behavioral graph remains under temporal drift and adversarial camouflage.

\subsection{Approach}
\label{sec:Approach}
Our framework operates by converting raw execution traces into compact behavioral patterns and analyzing them through information-theoretic distances. We first define a unified sequence modeling method to express multi-dimensional transaction properties as a structural script language. This representation is then used to construct an implicit behavioral graph and filter out global tracking bias, allowing for parameter-free malicious coordination discovery.

\subsubsection{Transaction Grammar and Sequence Encodings}
\label{sec:transaction_grammar}

To capture on-chain behavior compactly, each transaction is encoded as \(x_t = (\tau_t, \gamma_t, \nu_t)\): Rhythm, Structure, and Intent.

\begin{itemize}
    \item \textbf{Rhythm (\(\tau_t\))}: This component captures the temporal patterns of activity, ranging from high-frequency machine-speed bursts to human scales of dormancy. We compute the inter-arrival time \(\Delta t\) in seconds since the previous transaction and discretize it into 16 log-bins, ranging from \texttt{same-block} (\(\Delta t = 0\)) to \texttt{dormancy} (\(\Delta t > 30\) days), with a unique \texttt{START} token for the beginning of a sequence.

    \item \textbf{Structure (\(\gamma_t\))}: To represent the inner workings of each transaction, we define nine properties from the EVM trace, including call depth, branching, and error rates. These metrics are discretized and concatenated into a unique composite string (e.g., \texttt{Cnt:2\_Dp:4\_Err:0}), effectively treating the ``shape'' of the execution trace as a word in our vocabulary.

    \item \textbf{Intent ($\nu_t$)}: This element encodes semantic intent through the 4-byte function selector of the call. In order to keep control of vocabulary size, the primary experiment uses a vocabulary of \(|V|=100\) most common selectors on a class-stratified basis, and less common selectors are treated as the generic \texttt{[UNK]}. We additionally verify that an unsupervised global-frequency vocabulary of the same size yields equivalent NCD retrieval quality, showing that the construction does not materially depend on class labels.
\end{itemize}

The grammar supports multiple sequence types: \texttt{rhythm\_only}, \texttt{coarse\_intent}, and \texttt{full\_tokens}. The 16 logarithmic rhythm bins preserve activity scales from same-block bursts to long dormancy, while the \(|V|=100\) selector vocabulary keeps the symbolic alphabet compact enough for compression.

\begin{table}[htbp]
\caption{Transaction Grammar: Token Vocabulary Formulation}
\label{tab:token_vocab}
\centering
\begin{tabularx}{\textwidth}{@{}lX@{}}
\toprule
\textbf{Token Type} & \textbf{Definition and Discretization Strategy} \\
\midrule
Rhythm (\(\tau\)) & Log-binned \(\Delta t\) (16 bins): \(\{\texttt{START}, \texttt{Same-Block}, \dots, \texttt{30d+}\}\). \\
Structure (\(\gamma\)) & Composite string of 8 binned trace metrics: Complexity (Count, Depth, Branching), Outcome (Reverts, Root status), Op Types (Delegate, Static, Create ops). \\
Intent (\(\nu\)) & Compact vocabulary of \(|V|=100\) frequent function selectors; global-frequency variant tested as ablation. Special: \texttt{[UNK]}, \texttt{NO\_CALLDATA}. \\
\bottomrule
\end{tabularx}
\end{table}

\subsubsection{Normalized Compression Distance}
\label{sec:ncd}

We quantify sequence similarity using Normalized Compression Distance (NCD) \cite{cilibrasi2004clusteringcompression}. We use gzip as a deterministic, widely available LZ-family compressor that provides a reproducible baseline for detecting repeated symbolic motifs; alternative compressors are left for future work. For any two encoded wallet sequences \(x\) and \(y\), the distance is defined as:
\begin{equation}
\mathrm{NCD}(x,y) = \frac{C(xy) - \min(C(x), C(y))}{\max(C(x), C(y))}
\end{equation}
where \(C(x)\) represents the byte length of the sequence \(x\) after compression with \texttt{gzip}, and \(xy\) denotes the concatenation of the two sequences. The NCD value ranges from 0 to 1, where lower values indicate higher structural similarity.

The efficacy of NCD in this context relies on the distinction between algorithmic regularity and human entropy. In our domain, organic users exhibit high natural entropy and diverse transaction grammars. Conversely, Sybil farmers rely on automated scripts that produce highly redundant sequences. When two such scripted sequences are concatenated (\(xy\)), the compressor identifies the shared structural motifs and achieves a higher compression ratio than it would for disparate organic traces, thereby yielding a low NCD score. This allows NCD to act as a coordination signal that is inherently robust to the exact timing or value of transactions, focusing instead on the shared ``logic'' of execution.

\subsubsection{Implicit Behavioral Graph Construction}
\label{sec:implicit_graph}

While traditional blockchain analytics focuses on explicit interaction graphs (i.e., value flow), we construct an \textit{implicit behavioral graph} derived from the pairwise similarity of transaction grammars. For a given sample of \(N = 900\) wallets, we compute a symmetric \(N \times N\) distance matrix where each element \(D_{ij}\) is the NCD between wallet \(i\) and wallet \(j\). We transform this distance into a similarity score \(s = 1 - \mathrm{NCD}\) to serve as the adjacency weight for our graph.

This implicit graph representation serves three primary downstream purposes:

\begin{itemize}
    \item \textbf{Local Neighborhood Retrieval:} By examining the \(k\)-nearest neighbors (\(k\)-NN) of a suspect wallet in the NCD space, we can identify other actors within the same coordinated campaign. Our empirical results (H3) suggest that these local neighborhoods are significantly enriched with Sybil candidates, even when the actors have no direct on-chain funding links.

    \item \textbf{Global Community Discovery:} The similarity matrix serves as a primitive for unsupervised partitioning. We apply Spectral and Agglomerative clustering to detect larger pockets of coordination. Unlike isolated classification, this approach can identify emerging Sybil strategies by grouping wallets that exhibit similar ``fingerprints'' of structural regularity.

    \item \textbf{Robustness Analysis:} Treating similarity as a graph allows us to evaluate the structural resilience of the signal. By re-evaluating the graph connectivity before and after the application of the Blind-Spot protocol, we can quantify how much of the ``coordination'' is driven by shared utility (label leakage) versus the underlying algorithmic logic of the scripts.
\end{itemize}

By reinterpreting pairwise compression as a graph structure, we move from the classification of individual entities to the discovery of functional communities, making the framework particularly suitable for the open-world discovery of ``orphaned'' Sybil clusters.

\subsubsection{Blind-Spot Protocol for Leakage-Free Evaluation}
\label{sec:blind_spot}

For leakage-controlled evaluation, our label-informed Blind-Spot protocol removes high-signal counterparties (e.g., popular DeFi protocols) that introduce label leakage by acting as shortcut features \cite{LeakageDataMining,Geirhos_2020}. For instance, interactions with the Uniswap Router are a strong circular indicator of MEV arbitrage, while the ENS Controller is a primary signal for human-driven Organic identity.

The protocol systematically identifies and removes these high-signal interactions to force models to rely on authentic behavioral signatures, such as transaction timing and execution traces. In our implementation, this process resulted in the removal of \(25.5\%\) of the total transaction volume, targeting ubiquitous or class-specific contracts like OpenSea Wyvern and WETH9. By mitigating the influence of these specific counterparties, the protocol ensures that the predictive signal is derived from the structural regularity of the actor's behavior rather than superficial interaction targets.

\section{Experiments and Results}

This section evaluates the proposed compression-based behavioral similarity framework under both Raw and Blind-Spot-filtered settings. We first describe dataset construction and leakage filtering, then test behavioral separation, local candidate discovery, temporal stability, and robustness under synthetic camouflage.

\subsection{Experimental Setup and Dataset}
Our dataset was built by fetching the labels and addresses from two major sources. Organic and Sybil actors have been sourced from the Hop Protocol airdrop list \cite{hop_airdrop_github_2026}. The labels for the MEV Bots have been collected using the labeled datasets from Dune Analytics with a hard cutoff at May 13th, 2022, when Hop Protocol released its snapshot. We acquired addresses with labels indicating Arbitrage Bots. For downloading transaction's internal execution traces, we leveraged Google BigQuery. We initially excluded any wallet with less than $L_{\min}=10$ transactions.

To refine this raw dataset and prevent label leakage, we applied our Blind-Spot Protocol. In the scope of this protocol, we calculated three key measures for each candidate address:
\begin{itemize}
    \item \textbf{Support ($n$)}: the total number of transactions related to an address. A minimum threshold of $n \ge 50$ is applied to ensure statistical relevance.
    \item \textbf{Max Class Share (Purity $P$)}: proportion of transactions that belong to the predominant actor class. Address with Purity value higher than $P \ge 0.95$ may be considered as leaking one's class.
    \item \textbf{Shannon Entropy ($H$)}: uncertainty of class distribution measured as $H = -\sum p_i \log_2(p_i)$ . Leaking addresses have entropy close to zero.
\end{itemize}

All transactions, including interactions with these ``blind-spots`` addresses, are removed from the histories of all wallets in the dataset. Applying this protocol removed 25.5\% of all transaction volume and excluded 443 wallets that dropped below the minimum-history threshold ($L_{\min}=10$). The final dataset statistics, comparing the raw and leakage-free wallet counts, are summarized in Table~\ref{tab:dataset_stats_final}. For runtime context, computing a full \(N=900\) NCD matrix over 404,550 wallet pairs required 83--96 seconds across repeated runs, with a mean of 90.4 seconds, on a 2020 MacBook Pro with Apple M1 and 16 GB RAM.

\begin{table}[H]
\caption{Dataset Statistics}
\label{tab:dataset_stats_final}
\centering
\begin{tabularx}{\textwidth}{@{}X r@{\hspace{1.8em}}r@{}}
\toprule
\textbf{Class}
& \begin{tabular}[c]{@{}r@{}}\textbf{Raw}\\\textbf{Wallets Count}\end{tabular}
& \begin{tabular}[c]{@{}r@{}}\textbf{Leakage-Free}\\\textbf{Wallets Count}\end{tabular} \\
\midrule
Organic & 9,282 & 9,204 \\
MEV Bot & 3,558 & 3,202 \\
Sybil   & 1,764 & 1,755 \\
\midrule
\textbf{Total} & \textbf{14,604} & \textbf{14,161} \\
\bottomrule
\end{tabularx}
\end{table}

\subsection{Statistical Analysis of Behavioral Similarity}
\label{sec:behavioral_similarity}

To test whether Sybil wallets are more behaviorally similar than Organic and MEV wallets, and whether this signal survives leakage control, we compare within-class NCD similarity before and after Blind-Spot filtering. Behavioral coordination is measured via within-class pairwise similarity
\(s = 1 - \mathrm{NCD}\) using \(N = 10\) random seeds
(300 wallets per class, 44,850 pairs per class and seed). The separation of clusters is
assessed through the mean similarity difference \(\Delta\) between classes and
Cliff's \(\delta\) statistic, defined as:
\begin{equation}
\delta = \Pr(X > Y) - \Pr(X < Y).
\end{equation}
Here, \(X\) and \(Y\) are two distributions of pairwise similarities. This effect-size statistic is suitable for non-normal data. Standard pairwise inference assumes independent observations, but each wallet appears in multiple comparisons. Therefore,
we treat \(\delta\) as a purely descriptive statistic and assess significance by
testing whether the ten seed-level \(\Delta\) values are significantly greater
than zero using a one-sided Wilcoxon signed-rank test.

\begin{table}[H]
\caption{Within-class similarity differences (\(\Delta\)) and Cliff's
\(\delta\) effect sizes. Values are means \(\pm\) standard deviations across
\(N = 10\) random seeds.}
\label{tab:h1h2_results}
\centering
\small
\renewcommand{\arraystretch}{1.15}
\setlength{\tabcolsep}{4pt}
\begin{tabularx}{\textwidth}{@{}l l c c c c@{}}
\toprule
\textbf{Encoding} &
\textbf{Setting} &
\textbf{Syb--Org \(\Delta\)} &
\textbf{Cliff's \(\delta_{\mathrm{Org}}\)} &
\textbf{Syb--MEV \(\Delta\)} &
\textbf{Cliff's \(\delta_{\mathrm{MEV}}\)} \\
\midrule
\texttt{rhythm\_only}  & Raw
& \(0.089 \pm 0.008\)  & \(0.392\)
& \(0.092 \pm 0.009\)  & \(0.404\) \\

\texttt{rhythm\_only}  & LF
& \(0.115 \pm 0.010\)  & \(0.483\)
& \(0.117 \pm 0.011\)  & \(0.490\) \\
\midrule

\texttt{rhythm+intent} & Raw
& \(0.122 \pm 0.011\)  & \(0.384\)
& \(0.118 \pm 0.010\)  & \(0.368\) \\

\texttt{rhythm+intent} & LF
& \(0.233 \pm 0.012\)  & \(0.674\)
& \(0.214 \pm 0.013\)  & \(0.640\) \\
\midrule

\texttt{full\_tokens}  & LF\textsuperscript{a}
& \(0.091 \pm 0.010\)  & \(0.445\)
& \(0.087 \pm 0.009\)  & \(0.428\) \\
\bottomrule
\end{tabularx}

\medskip
\footnotesize
LF denotes Leakage-Free.
\textsuperscript{a} We report \texttt{full\_tokens} only in the leakage-free
setting, where the complete transaction grammar is evaluated after removing
utility-based shortcut interactions. Raw-vs-leakage comparisons are therefore
limited to the first two encodings.
\end{table}

\begin{figure}
    \centering
    \includegraphics[width=0.65\linewidth]{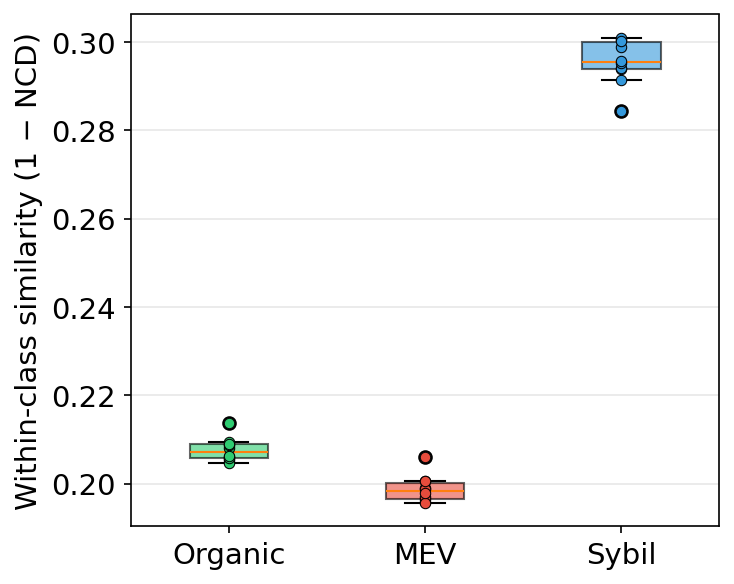}
    \caption{Distribution of within-class behavioral similarity (1-NCD) for the Full Tokens grammar. The Sybil class exhibits a markedly higher similarity baseline compared to Organic and MEV actors.}
    \label{fig:dist_full_tokens}
\end{figure}

Table~\ref{tab:h1h2_results} indicates that Sybil wallets show higher
within-class similarity than Organic and MEV wallets across all tested seeds. For the full\_tokens leakage-free representation, this distributional
shift is visualized in Fig.~\ref{fig:dist_full_tokens}.
For rhythm+intent profiles in the Leakage-Free setting, the observed
\(\delta = 0.674\) against Organic wallets corresponds to a common-language
effect size of approximately \(0.837\), excluding ties. Thus, Sybil--Sybil pairs are more likely to be more similar than Organic--Organic pairs.
This supports the hypothesis that behavioral scripts are reused more often in Sybil attacks.

When both Raw and Leakage-Free variants are available, the Blind-Spot protocol
yields a larger effect size. For the \texttt{rhythm+intent} encoding, the
Sybil--Organic similarity gap nearly doubles, increasing from
\(+0.122\) to \(+0.233\) after the exclusion of high-signal global contracts.
We interpret this as a denoising effect: removing universal contract
counterparties, such as Uniswap, reduces shared ``utility'' similarity and
reveals a clearer signal of repetitive script structure.

\subsection{Candidate Discovery and Comparative Analysis}
\label{sec:candidate_discovery}

To test whether compression-based similarity can support Sybil candidate discovery without financial links, we evaluate nearest-neighbor retrieval in the NCD behavioral graph and compare it with symbolic and supervised baselines. Unlike explicit financial graphs that require direct token transfers, this graph connects wallets based on structural regularity. To determine whether this similarity effectively isolates malicious actors, we analyze local neighborhood retrieval using the class prior of $0.333$ as a baseline. As shown in Table~\ref{tab:retrieval_results}, the \texttt{full\_tokens} representation using NCD outperforms the symbolic baseline in mean performance, achieving a 1-NN accuracy of $0.696 \pm 0.030$. We compare NCD against a TF-IDF cosine similarity baseline, which treats the transaction grammar as a ``bag-of-symbols.'' This suggests NCD captures structure beyond token frequency, with a $5.1\%$ lead over TF-IDF.

The strength of the NCD signal is most evident in the neighborhood enrichment metrics. For the \texttt{full\_tokens} representation, the top-10 neighborhood purity reaches $0.754 \pm 0.019$. This indicates that for any given Sybil candidate, roughly $75\%$ of its immediate behavioral neighbors are also coordinated actors. Furthermore, the top-10 recall of $0.922 \pm 0.015$ demonstrates that nearly all Sybil wallets in the dataset are linked to at least one other member of a coordinated campaign within their immediate neighborhood. While global clustering metrics such as ARI and NMI remain moderate, typically around $0.10$, we observe that Sybil scripts form dense, high-purity pockets within the similarity graph. For discovery purposes, this local density is highly actionable, as it allows auditors to expand a single ``seed'' address into a complete coordinated cluster without requiring explicit on-chain funding links.

\begin{table}[htbp]
\caption{Comparative analysis of accuracy across methods and data settings ($N=10$ seeds). Accuracy is reported as mean $\pm$ standard deviation.}
\label{tab:h4_full_results}
\centering
\small
\begin{tabular}{@{}l l c c@{}}
\toprule
\textbf{Method} &
\textbf{Encoding} &
\begin{tabular}[c]{@{}c@{}}\textbf{Raw}\\\textbf{Acc.}\end{tabular} &
\begin{tabular}[c]{@{}c@{}}\textbf{Leakage-Free}\\\textbf{Acc.}\end{tabular} \\
\midrule
NCD 1-NN  & \texttt{rhythm\_only}   & $0.533 \pm 0.028$ & $0.521 \pm 0.008$ \\
XGBoost   & \texttt{rhythm\_only}   & $0.744 \pm 0.018$ & $0.748 \pm 0.017$ \\
TF-IDF+LR & \texttt{rhythm\_only}   & $0.641 \pm 0.016$ & $0.638 \pm 0.017$ \\
BiLSTM    & \texttt{rhythm\_only}   & $0.684 \pm 0.020$ & $0.681 \pm 0.019$ \\
\midrule
NCD 1-NN  & \texttt{rhythm+intent}  & $0.593 \pm 0.008$ & $0.560 \pm 0.018$ \\
XGBoost   & \texttt{rhythm+intent}  & $0.796 \pm 0.014$ & $0.801 \pm 0.013$ \\
TF-IDF+LR & \texttt{rhythm+intent}  & $0.688 \pm 0.015$ & $0.684 \pm 0.016$ \\
BiLSTM    & \texttt{rhythm+intent}  & $0.704 \pm 0.018$ & $0.701 \pm 0.017$ \\
\midrule
NCD 1-NN  & \texttt{full\_tokens}   & $0.713 \pm 0.024$ & $0.703 \pm 0.013$ \\
XGBoost   & \texttt{full\_tokens}   & $0.785 \pm 0.007$ & $0.795 \pm 0.007$ \\
TF-IDF+LR & \texttt{full\_tokens}   & $0.666 \pm 0.012$ & $0.663 \pm 0.016$ \\
BiLSTM    & \texttt{full\_tokens}   & $0.708 \pm 0.018$ & $0.703 \pm 0.019$ \\
\bottomrule
\end{tabular}
\end{table}
\FloatBarrier

To contextualize the training-free approach, we compare NCD against three supervised baselines: XGBoost \cite{Chen_2016}, TF-IDF with Logistic Regression \cite{manning2008introduction}, and a hierarchical BiLSTM \cite{Longshort}.

\begin{table}[t]
\caption{Neighborhood retrieval and baseline comparison ($N=10$ seeds). Accuracy and purity are reported as mean $\pm$ standard deviation. The class prior for Sybil wallets is $0.333$.}
\label{tab:retrieval_results}
\centering
\small
\begin{tabularx}{\textwidth}{@{}l l c c c c@{}}
\toprule
\textbf{Encoding} &
\textbf{Method} &
\begin{tabular}[c]{@{}c@{}}\textbf{1-NN}\\\textbf{Acc.}\end{tabular} &
\begin{tabular}[c]{@{}c@{}}\textbf{5-NN}\\\textbf{Acc.}\end{tabular} &
\begin{tabular}[c]{@{}c@{}}\textbf{Top-10}\\\textbf{Pur.}\end{tabular} &
\begin{tabular}[c]{@{}c@{}}\textbf{Top-10}\\\textbf{Rec.}\end{tabular} \\
\midrule
\texttt{rhythm\_only}   & NCD           & $0.529 \pm 0.025$ & $0.581 \pm 0.020$ & $0.560 \pm 0.022$ & $0.844 \pm 0.018$ \\
\texttt{rhythm\_only}   & TF-IDF cosine & $0.537 \pm 0.013$ & $0.559 \pm 0.018$ & $0.459 \pm 0.017$ & $0.858 \pm 0.023$ \\
\midrule
\texttt{rhythm+intent}  & NCD           & $0.565 \pm 0.014$ & $0.624 \pm 0.019$ & $0.627 \pm 0.026$ & $0.856 \pm 0.019$ \\
\texttt{rhythm+intent}  & TF-IDF cosine & $0.573 \pm 0.014$ & $0.588 \pm 0.014$ & $0.505 \pm 0.019$ & $0.813 \pm 0.020$ \\
\midrule
\texttt{full\_tokens}   & NCD           & $0.696 \pm 0.030$ & $0.728 \pm 0.018$ & $0.754 \pm 0.019$ & $0.922 \pm 0.015$ \\
\texttt{full\_tokens}   & TF-IDF cosine & $0.645 \pm 0.016$ & $0.671 \pm 0.017$ & $0.606 \pm 0.030$ & $0.836 \pm 0.025$ \\
\bottomrule
\end{tabularx}
\end{table}

As detailed in Table~\ref{tab:h4_full_results}, while the tree-based XGBoost model achieves the highest absolute accuracy of $0.795 \pm 0.007$, NCD remains highly competitive. Notably, in the leakage-free environment using the \texttt{full\_tokens} encoding, NCD achieves mean performance parity with the complex BiLSTM architecture, with both reaching $0.703$. This is a significant finding: a parameter-free, compression-based metric can match the performance of a deep learning model without requiring labeled training sets or the computational overhead of backpropagation.

Moreover, NCD shows robustness under varying data regimes and additional leakage and scaling evaluations. On \texttt{full\_tokens}, NCD performance differs by less than \(1\%\) between the Raw and Leakage-Free settings: \(0.713 \pm 0.024\) versus \(0.703 \pm 0.013\). This suggests that Blind-Spot filtering acts more as denoising than as a strict prerequisite.

To check whether label-driven vocabulary construction causes this effect, we replace the class-balanced Intent vocabulary with a purely unsupervised vocabulary of the same size, built from global frequency statistics (\(|V|=100\)). The 1-NN accuracy on full-token changes by at most \(0.003\). Across Blind-Spot elimination levels from \(0\%\) to \(54\%\), the top-10 Sybil recall remains stable, ranging from \(0.909\) to \(0.929\).

As a final experiment, a MinHash/LSH candidate generator reduces the \(N=900\) lookup time from approximately \(98\) s to approximately \(2\) s without degrading 1-NN accuracy: \(0.706\) versus \(0.704\) for exact lookup. It also improves top-10 recall, \(0.936\) versus \(0.925\), although top-10 purity decreases from \(0.730\) to \(0.654\). This suggests a possible two-stage architecture in which the first stage generates approximate candidates using MinHash/LSH and the second reranks them with exact NCD similarity.

We do not include a GNN baseline because graph neural methods over explicit transfer or funding graphs evaluate a different signal from the disconnected behavioral-similarity setting targeted here. Direct comparison with graph-based discovery systems remains future work.

\subsection{Temporal Split: Cross-Period Candidate Discovery}
\label{sec:temporal_split}

To evaluate whether the grammar generalizes across time rather than only within a single static snapshot, a temporal split experiment is conducted. First, transactions of each wallet are sorted in chronological order and divided at the midpoint into early and late windows; wallets are retained only if both windows contain at least 10 transactions after Blind-Spot filtering. We use the \texttt{full\_tokens} leakage-free setting with 300 wallets per class and three random seeds.

First, each late window is used as a query and early window sequences of all wallets are ranked via gzip-NCD without considering the wallet's own early window. This way, the experiment evaluates the ability to find other actors similar in behavior in different periods rather than trivial self-matches. Also, we conduct a leave-one-out evaluation within the late windows.

\begin{table}[H]
\caption{Temporal split retrieval. Late-vs-late is a within-period half-length control; late-to-early evaluates cross-period candidate discovery.}
\label{tab:temporal_split}
\centering
\small
\setlength{\tabcolsep}{6pt}
\begin{tabular}{@{}lccc@{}}
\toprule
\textbf{Setting} & \textbf{1-NN Acc.} & \textbf{Top-10 Purity} & \textbf{Top-10 Recall} \\
\midrule
Within-period & 0.693 & 0.707 & 0.933 \\
Cross-period & 0.619 & 0.693 & 0.782 \\
\bottomrule
\end{tabular}
\end{table}

Temporal transfer is observed with attenuation. While the cross-period 1-NN accuracy stays well above the class prior of 0.333 and the top-10 Sybil purity is not much lower than within-period controls, top-10 recall decreases from 0.933 to 0.782. This indicates real temporal drift: behavior of some Sybil wallets in the later period becomes unconnected to their earlier campaign behavior even within the top-10 neighborhood. So, the experiment shows temporal stability of the grammar within campaigns while cross-campaign transfer is still an open question for future work.

\subsection{Camouflage Sensitivity and Failure Boundary}
\label{sec:camouflage_sensitivity}

To test stability under adversarial camouflage, we perturb Sybil sequences and compare NCD retrieval with the strongest supervised baseline, XGBoost. We use the \texttt{full\_tokens} encoding because it contains the richest rhythm, execution-flow, and function-selector information.

We apply a synthetic blend-in stress test with composite noise level \(\alpha \in [0.0,\allowbreak 0.1,\allowbreak 0.2,\allowbreak 0.3,\allowbreak 0.4,\allowbreak 0.5]\). The perturbations are not calibrated from observed Ethereum Sybil campaigns; they approximate plausible evasion behaviors: assigning an \(\alpha\) fraction of transactions to neighboring rhythm bins, inserting an \(\alpha\)-proportional number of organic-looking chaff transactions, and permuting an \(\alpha\) fraction of adjacent transaction pairs.
We refer to this combined perturbation as the ``noise level'' $\alpha$, treating it as a controlled failure-boundary test rather than a verified model of real-world Sybil evasion.

\begin{table}[H]
\caption{Sybil recall versus composite noise level $\alpha$. Values represent the mean $\pm$ standard deviation across $N=5$ random seeds.}
\label{tab:camouflage_recall}
\centering
\setlength{\tabcolsep}{12pt} 
\small
\begin{tabular}{@{}l c c@{}}
\toprule
\textbf{Noise Level $\alpha$} &
\textbf{NCD 1-NN} &
\textbf{XGBoost} \\
\midrule
$0\%$  & $1.000 \pm 0.000$ & $1.000 \pm 0.000$ \\
$10\%$      & $1.000 \pm 0.000$ & $0.973 \pm 0.013$ \\
$20\%$      & $0.999 \pm 0.002$ & $0.927 \pm 0.021$ \\
$30\%$      & $0.997 \pm 0.002$ & $0.891 \pm 0.042$ \\
$40\%$      & $0.991 \pm 0.005$ & $0.845 \pm 0.054$ \\
$50\%$      & $0.981 \pm 0.011$ & $0.816 \pm 0.062$ \\
\bottomrule
\end{tabular}
\end{table}

The findings, based on $N=5$ random seeds and presented in Table~\ref{tab:camouflage_recall}, indicate the striking disparity in resilience to this attack between XGBoost and NCD 1-NN. The accuracy degrades in a linear manner depending on the increase in the ``noise level''. The brittleness in this case can be explained by the fact that the algorithm makes an implicit assumption about the distribution of features used to train the model (token counts and entropy scores), which can be easily violated by adding organic-looking junk data. On the contrary, the recall of NCD 1-NN shows much better resistance to camouflaging, achieving $0.981$ recall even under the maximal ``noise level'' of $50\%$. Since the gzip-based distance detects shared motifs and sub-sequences of transactions, the basic ``algorithmic DNA'' of the Sybil script can be preserved under these modifications.

\section{Discussion}
\label{sec:discussion}

Our results validate that compression-based similarity serves as an effective local primitive for candidate expansion. Instead of being a robust general-purpose detector, NCD is more applicable in the discovery of areas where there is some kind of structural regularity that may indicate a collaboration between Sybils. This collaboration becomes evident in the high Top-10 neighborhood recall score (\(0.922\)) generated by the \texttt{full\_tokens} representation, indicating that similar wallets could be connected even as orphans in traditional financial interaction graphs.

Crucially, NCD proves highly resilient to synthetic perturbations. Unlike XGBoost, which degrades as noise increases, NCD maintains a recall of (\(0.981\) for \(\alpha = 50\%\) threshold). This can be explained by the dictionary nature of the \texttt{gzip} compression method, which can recognize motifs even if some artificial ``chaff'' is introduced to confuse the pattern recognition system, along with time perturbation and shuffling of transactions locally. However, these results should be interpreted as robustness under synthetic stress, not as evidence that the attacks fully reproduce observed Ethereum Sybil evasion.

The Blind-Spot protocol plays the role of a leakage-control denoiser that moves the similarity information from the utility signal to transactional-grammatical signals. The experiments showed that there is an important distinction in the representations we used: although the representation that combines \texttt{rhythm+intent} has a larger separation gap between Sybils and Organics, the \texttt{full\_tokens} grammar performs better on the local search task. We believe that this can be attributed to the difference in granularity: rhythms and intents encode class distinctions, while the EVM traces provide the necessary detail for accurate local searches.

Some limitations remain. First, the Blind-Spot protocol is label-informed and serves as a leakage-control evaluation tool rather than a fully label-free deployment procedure. Second, NCD avoids model retraining but requires \(O(N^2)\) pairwise distances for global graph construction: scaling from \(N=900\) to \(N=10{,}000\) increases the matrix from 404,550 to 49,995,000 distances, implying about 3.1 hours under the same implementation and average sequence length. However, local candidate expansion from a known suspicious seed wallet scales linearly in the candidate universe, making targeted audits more tractable. In summary, the present evaluation demonstrates training-free local candidate discovery within the Hop-centered dataset and provides evidence of within-campaign temporal stability. However, cross-campaign transfer, deployment on a second airdrop or governance campaign, and unknown-class rejection remain outside the scope of the current evidence.

\section{Future Work}
\label{sec:future_work}

We propose analyzing the difference between gzip compression and other compression algorithms, such as Brotli, Zstandard, and LZMA, and studying how they differ in representing various types of repetition in transaction grammars. Earlier literature about NCD applications for malware analysis has shown the significant impact of such factors as compressor choice, object size, and regularity structure on similarity \cite{alshahwan2015detectingmalwareinformationcomplexity}. A possible extension would include comparison of NCD with learned Transformer-based sequence models and ranking objectives from financial time-series and cryptocurrency literature \cite{szydlowski2024hidformer,kwiatkowski2025lossfunctions,tokajuk2025partialmultivariatetransformertool}.

Although the temporal split provides evidence of within-campaign stability, future work should test cross-campaign transfer using a second airdrop, governance vote, or another EVM-compatible environment. This would separate stability within one campaign from generalization across different Sybil strategies. For production-scale global discovery, our MinHash/LSH results suggest a two-stage architecture: approximate candidate generation \cite{ApproximateNearestNeighbor,ProceedingsCompressionandComplexity} followed by exact NCD reranking, avoiding exhaustive all-pairs compression over the full wallet universe.

Finally, future work should evaluate the methodology in broader deployment scenarios, including later Ethereum activity periods, Layer-2 and other EVM-compatible chains, and adaptive camouflage strategies.

\section{Conclusion}
\label{sec:conclusion}

The Sybil attack represents a persistent threat for blockchain systems, posing as regular participants to steal airdrops and abuse decentralized governance. In this paper, we tackled the problems arising from the reliance on closed-set models by presenting a compression-based approach to behavioral similarity. Through the use of EVM execution traces encoded into symbolic Transaction Grammars together with the implementation of a Blind-Spot leakage-prevention protocol, we have shown that even a highly complex machine learning model, such as LSTM networks, is not necessary for identifying behavioral similarity. In particular, we presented how a simple NCD model, which takes advantage of the redundancies in automated wallet scripts, can reach the same mean accuracy score as our hierarchical BiLSTM benchmark without the need for model training.

The most important contribution is the development of a behavior-based Sybil matching method that extends the current arsenal of techniques used for blockchain audits. Using NCD scores to construct implicit behavioral connections, the method can discover ``orphaned'' Sybil candidates that lack explicit financial links to known groups but share similar algorithmic structure. The temporal split further shows that this signal has within-campaign stability: late-window behavior still retrieves Sybil-dominated early-window neighborhoods, although with reduced top-10 recall. Finally, under controlled synthetic camouflage stress tests, NCD remains substantially more stable than the supervised XGBoost baseline.

In conclusion, our framework supports Sybil audits by moving from passive label-based classification to proactive candidate discovery based on behavioral patterns. The community could take advantage of our work to use different types of data compression techniques, such as Brotli or Zstandard, to identify different execution patterns. Ultimately, application of the established primitives in other layer-2 environments, especially those involving human audits, will be the next important step towards securing decentralized systems.

\begin{credits}\subsubsection{\ackname}Large language models were used in this paper to improve grammar and language clarity. The authors have no competing interests to declare that are relevant to the content of this article

\end{credits}

\bibliographystyle{splncs04}
\bibliography{refs}

\end{document}